\documentclass[letterpaper]{article} 
\usepackage{aaai23}  
\usepackage{times}  
\usepackage{helvet}  
\usepackage{courier}  
\usepackage[hyphens]{url}  
\usepackage{graphicx} 
\usepackage{natbib}  
\usepackage{caption} 
\usepackage{algorithm}
\usepackage{algorithmic}

\usepackage{newfloat}
\usepackage{listings}
\DeclareCaptionStyle{ruled}{labelfont=normalfont,labelsep=colon,strut=off} 
\floatstyle{ruled}
\newfloat{listing}{tb}{lst}{}
\floatname{listing}{Listing}

\usepackage{multirow}
\usepackage{amsmath}
\usepackage{amssymb}
\usepackage{xspace}
\usepackage{booktabs}
\usepackage{paralist}

\newcommand{\KL}{\operatorname{D_{KL}}\xspace}
\newcommand{\RKL}{\operatorname{D_{RKL}}\xspace}

\newcommand{\LatEmb}{\ensuremath{H}\xspace}
\newcommand{\HLV}{\ensuremath{L}\xspace}
\newcommand{\CxtMem}{\ensuremath{M}\xspace}

\newtheorem{theorem}{Theorem}
\newtheorem{proposition}{Proposition}

\title{Towards Diverse, Relevant and Coherent Open-Domain Dialogue Generation via Hybrid Latent Variables}
\author {
    Bin Sun\textsuperscript{\rm 1}, 
    Yitong Li\textsuperscript{\rm 2,3}, 
    Fei Mi\textsuperscript{\rm 2}, 
    Weichao Wang\textsuperscript{\rm 2}, 
    Yiwei Li\textsuperscript{\rm 1}, 
    Kan Li\textsuperscript{\rm 1} 
}
\affiliations {
    \textsuperscript{\rm 1} School of Computer Science \& Technology, Beijing Institute of Technology\\
    \textsuperscript{\rm 2} Huawei Noah's Ark Lab \quad
    \textsuperscript{\rm 3} Huawei Technologies Ltd.\\
    \{binsun,liyiwei,likan\}@bit.edu.cn, \{liyitong3,mifei2,wangweichao9\}@huawei.com
}

\begin{document}

\maketitle

\begin{abstract}


Conditional variational models, using either continuous or discrete latent variables, are powerful for open-domain dialogue response generation.
However, previous works show that continuous latent variables tend to reduce the coherence of generated responses. In this paper, we also found that discrete latent variables have difficulty capturing more diverse expressions.
To tackle these problems, we combine the merits of both continuous and discrete latent variables and propose a \textbf{H}ybrid \textbf{L}atent \textbf{V}ariable (HLV) method.
Specifically, HLV constrains the global semantics of responses through discrete latent variables and enriches responses with continuous latent variables. Thus, we diversify the generated responses while maintaining relevance and coherence.
In addition, we propose \textbf{C}onditional \textbf{H}ybrid \textbf{V}ariational \textbf{T}ransformer (CHVT) to construct and to utilize HLV with transformers for dialogue generation.
Through fine-grained symbolic-level semantic information and \textit{additive Gaussian mixing}, we construct the distribution of continuous variables, prompting the generation of diverse expressions. Meanwhile, to maintain the relevance and coherence, the discrete latent variable is optimized by \textit{self-separation training}.
Experimental results on two dialogue generation datasets (\texttt{DailyDialog} and \texttt{Opensubtitles}) show that CHVT is superior to traditional transformer-based variational mechanism w.r.t. diversity, relevance and coherence metrics.
Moreover, we also demonstrate the benefit of applying HLV to fine-tuning two pre-trained dialogue models (PLATO and BART-base).\footnote{A full version of this paper will be available soon.}
\end{abstract}


\section{Introduction}

In recent years, great efforts have been devoted to the task of response generation in open domain dialogue. However, due to the one-to-many and many-to-one phenomena \citep{Seq2Seq-Sutskever-2014,FilteringData-Csaky-2019}, generative dialogue models often failed to generate diverse, relevant and coherent responses. Many existing methods \citep{VaeTextGeneration-Bowman-2016,kgCVAE-ZhaoTiancheng-2017,HVaeMN-ChenHongshen-2018,DialogWAE-GuXiaodong-2019,discrete-cvae-2019-emnlp,PLATOv1,GVTandSVT2020,Binsun-Sepacvae-ACL2021,DBLP:conf/cikm/WangGFCW21,DialogVED-2022-ACL-weichen} introduce latent variables to alleviate these problems.

\begin{figure}[t]
\centering
\includegraphics[width=0.45\textwidth]{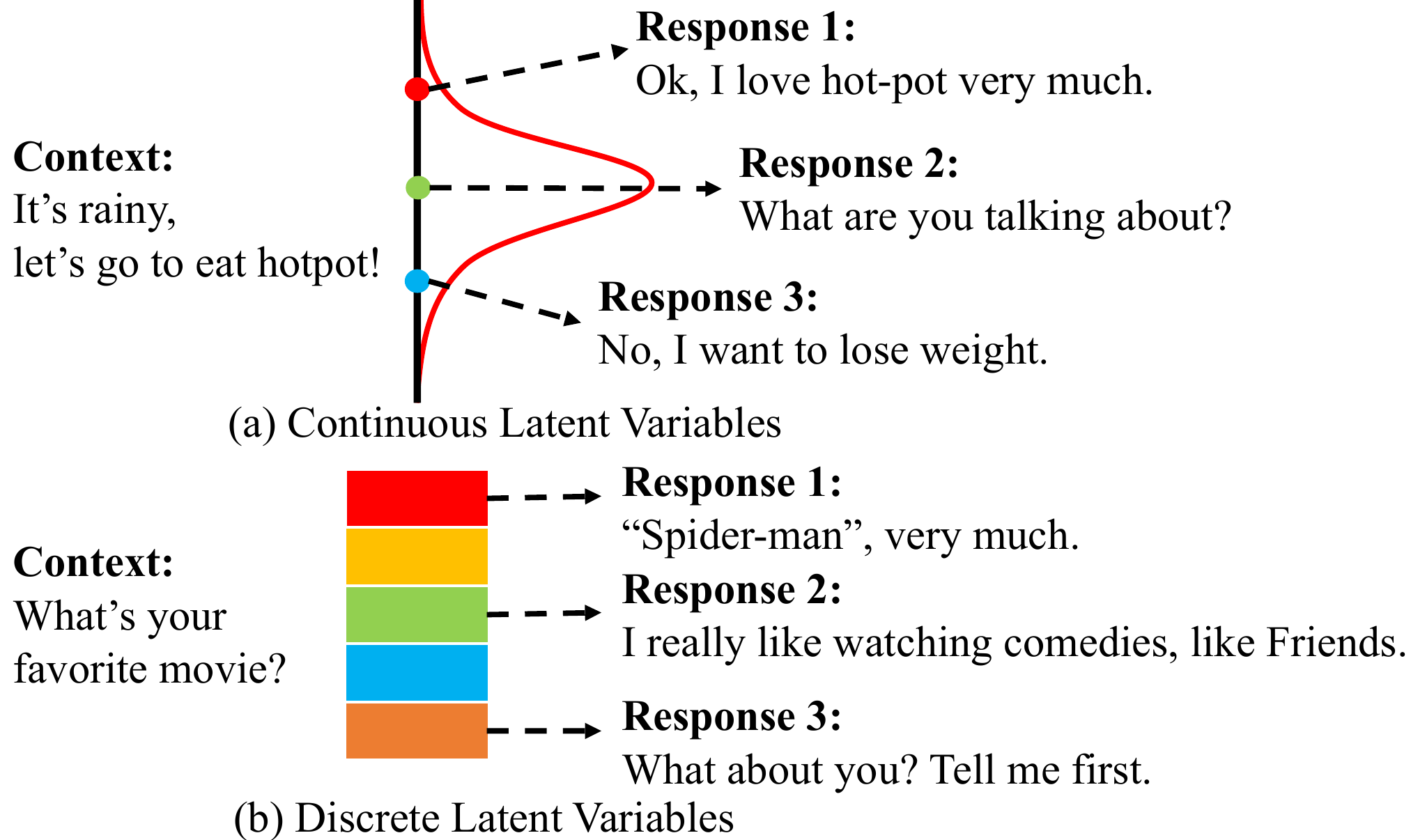} 
\caption{A schematic representation of the continuous and discrete latent variables.}
\label{fig1}
\end{figure}

As shown in Figure~\ref{fig1}, latent variables can be categorized into continuous latent variables and discrete latent variables from the perspective of their construction methods.
Continuous latent variable models \citep{VaeTextGeneration-Bowman-2016,kgCVAE-ZhaoTiancheng-2017,HVaeMN-ChenHongshen-2018,DialogWAE-GuXiaodong-2019,DBLP:conf/emnlp/WangFWZ19,Binsun-Sepacvae-ACL2021,DialogVED-2022-ACL-weichen} encode dialogue contexts and responses into a continuous latent space to capture diverse semantic features of dialogue responses.
In contrast, discrete latent variable models \citep{discrete-cvae-2019-emnlp,PLATOv1,DBLP:conf/acl/BaoHWWWWGLX21-PLATO-2} design a fixed number of learnable parameters to better capture the semantic relationship between responses and contexts.
However, continuous latent variables often face poor semantic correlation problem, where latent variables may not accurately reflect the semantics of the dialogue context \citep{SpaceFusion-GaoXiang-2019,Binsun-Sepacvae-ACL2021}. 
While discrete latent variables are difficult to capture fine-grained and diverse expressions due to the limitation of their variable scale (demonstrated in \S\ref{sec:ablation}).


To tackle the aforementioned problems, we propose to construct Hybrid Latent Variables (HLV).
HLV employs discrete latent variables to constrain the semantic relevance and uses the diverse features learned by continuous latent variables to enrich the expressions of generated responses.
To capture richer semantic information, we compliment the commonly used sentence-level continuous latent variables constructed by $[CLS]$ \citep{GVTandSVT2020} with multiple token-level representations.
More specifically, we first create symbolic-level Gaussian probability distribution for all context tokens, and then we construct sentence-level Gaussian probability distribution through an \textit{additive Gaussian mixing} method \citep{DBLP:conf/nips/0009SL17-additiveGaussian}. 
To build discrete latent variables, we introduce a \textit{self-separation training} approach \citep{Binsun-Sepacvae-ACL2021}, which expands the input with trainable discrete latent variables.

Furthermore, we propose a Conditional Hybrid Vraiational Transformer (CHVT), which constructs HLV based on a basic transformer encoder-decoder framework \citep{Transformer-Vaswani-2017}.
Through theoretically analyzing the difficulty of optimizing conditional variational models for dialogue response generation, we propose two techniques to effectively train HLV and CHVT.
First, to alleviate the posterior vanishing problem caused by one-to-many samples, we propose to remove the Bag-of-word loss and set a large warm-up steps to annealing the KL divergence.
Second, we give a theoretical upper bound for the KL divergence under the condition that the expectation of the posterior distribution is non-negative. Hence, we propose a relaxed KL divergence to further mitigate the posterior vanishing problem during the optimization process.

We conduct extensive experiments on \texttt{DailyDialog} \citep{dailydialog2017} and \texttt{Opensubtitles} \citep{opensubtitles2016} datasets. Empirical results show that CHVT is superior to existing transformer-based latent variable mechanisms w.r.t. diversity, relevance and coherence metrics. To explore the effect of HLV on pre-trained models \citep{PLATOv1,DBLP:conf/acl/LewisLGGMLSZ20-BART,DBLP:conf/acl/BaoHWWWWGLX21-PLATO-2,PanGuBot-feimi-2022}, we also extend HLV to the fine-tuning process of PLATO and BART, and we find that incorporating HLV can help the model perform better on dialogue generation task.
Furthermore, we validate the advantage of the proposed HLV over continuous and discrete latent variables through ablation studies.
Our contributions are as follow:
\begin{compactitem}
  \item We propose the Conditional Hybrid Variational Transformer (CHVT), which employs the proposed Hybrid Latent Variables (HLV) to generate diverse, relevant and coherent dialogue responses.
  \item We theoretical prove the main problems of optimizing continuous latent variables, and we put together several tricks (removing the BOW loss, a KL annealing trick, a relaxed KL divergence) to solve them.
  \item We show extensive empirical results on \texttt{DailyDialog} and \texttt{Opensubtitiles} datasets to illustrate the superior performance of CHVT and the advantages of HLV in generating diverse, relevant and coherent responses.
\end{compactitem}

\section{Related Work}

Previous dialogue generation models usually generate short, dull and general responses \citep{Seq2Seq-Sutskever-2014,Seq2Seq-Sordoni-2015,Seq2Seq-ShangLifeng-2015}.
To tackle these problem, continuous latent variables are introduced into the dialogue generation models \citep{CVAE(SPhred)-ShenXiaoyu-2017,kgCVAE-ZhaoTiancheng-2017,HVaeMN-ChenHongshen-2018,DialogWAE-GuXiaodong-2019,GVTandSVT2020,TCVAE2021-LeFang,Binsun-Sepacvae-ACL2021,DialogVED-2022-ACL-weichen,SegCVAE-binsun-emnlp-2022}.
These models employ the conditional variational mechanism \citep{StochasticGradientVariationalBayes-kingma-2014,ELBO1-Sohn-2015,ELBO2-Yan-2016,VaeTextGeneration-Bowman-2016}, which estimates the posterior probability distributions $p(z|c,r)$ and the prior probability distribution $p(z|c)$ of latent variable $z$ based on the dialogue corpora, where $c$ denotes the context, $r$ denotes the response, and a context and a response together constitute a single-turn dialogue pair. During training, these models sample the continuous latent variable $z$ from $p(z|c,r)$ and maximize the conditional probability $p(r|c,z)$ to encode context and response information into the latent space. Meanwhile, they also minimize the KL-divergence $\KL(p(z|c,r)||p(z|c))$ to bring the two distributions closer together, thus constraining the continuous latent variables $z$ sampled from the prior distribution $p(z|c)$ for inference.

With continuous latent variables, dialogue models effectively ensure diversity of generated responses. However, due to the \textit{one-to-many} and \textit{many-to-one} phenomena, the continuous latent variables often fail to capture the correct contextual semantics, resulting in irrelevant and incoherent responses \citep{Binsun-Sepacvae-ACL2021}. 
Different from the continuous latent variables, discrete latent variables are better at producing relevant and coherent responses. For instance, \citet{discrete-cvae-2019-emnlp} uses discrete latent variables with explicit semantics to generate responses, making responses easily contain the semantic correlation. \citet{PLATOv1} uses latent act recognition to build the relationship between discrete latent variables and multiple responses, and proposes response selection to chose the generated responses of most coherent with the context. However, due to the limited scale, discrete latent variables may capture less diverse features than continuous latent variables.

In a word, to make full use of the advantages of continuous and discrete latent variables, we propose the HLV, which uses discrete latent variables to constrain the contextual semantic correlation, and employs continuous latent variables to capture the symbolic-level features for enriching diversity.

\section{Methodology}
We firstly introduce the preliminaries of conditional variational models, i.e. autoencoder and transformer based methods,
then illustrate our proposed Conditional Hybrid Variational Transformer (CHVT) model, and finally analyze the factors affecting conditional variational transformer through theoretical deduction followed with relevant solutions.

\subsection{Preliminaries}
\label{sec:preliminary}
\begin{figure*}[t]
\centering
\includegraphics[width=0.85\textwidth]{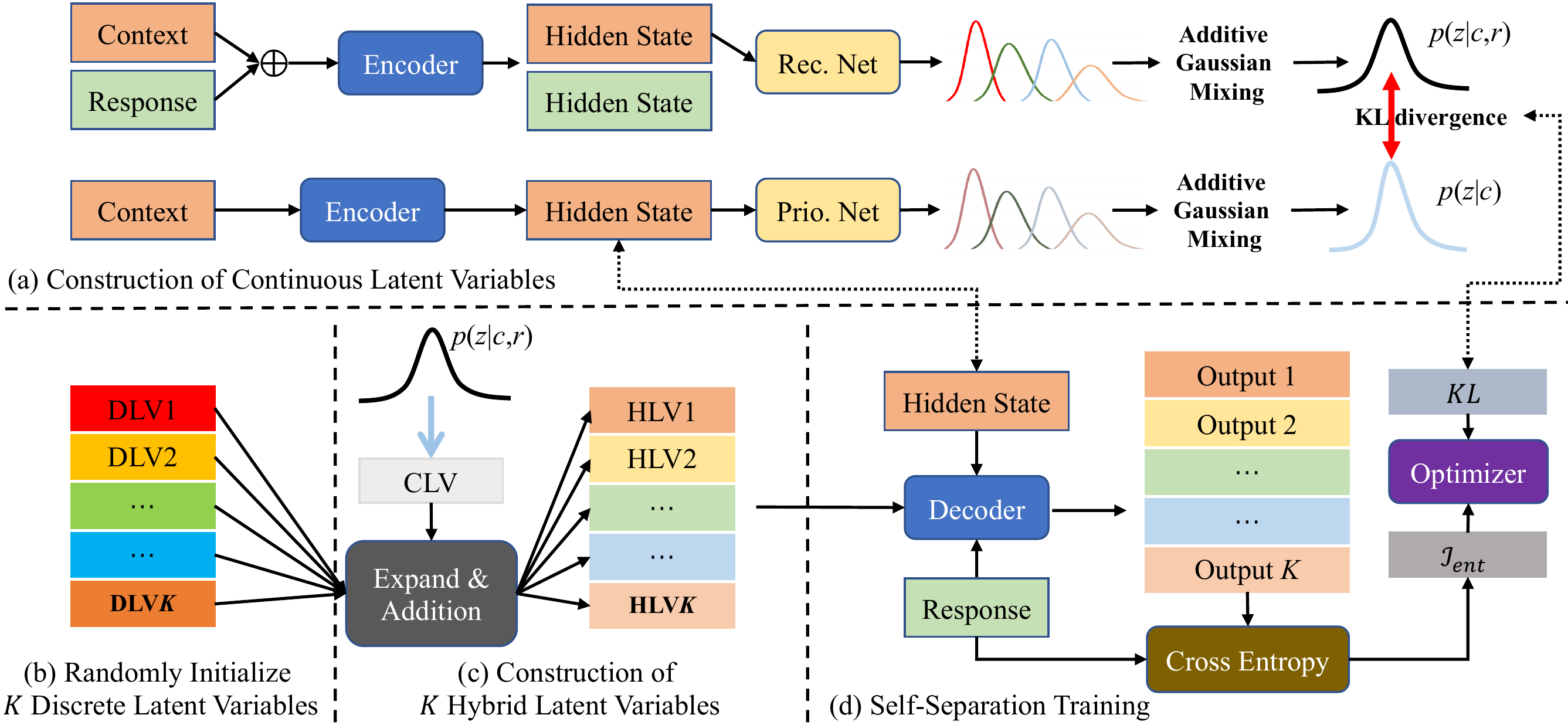}
\caption{The training steps of CHVT: (a) Constructing the continuous latent variables through symbolic-level Gaussian probability distributions and \textit{additive Gaussian mixing}; (b) Randomly initializing $K$ trainable discrete latent variables; (c) Building $K$ hybrid latent variables; (d) Optimizing CHVT with \textit{self-separation training}.}
\label{fig2}
\end{figure*}

\paragraph{Conditional Variational AutoEncoder}
In dialogue response generation task, Conditional Variational AutoEncoder (CVAE) is a powerful tool to ensure diversity and informativeness of the generated responses. 
CVAE primarily consists of three components: prior network, recognition network and generation network. The prior network is responsible for estimating the prior probability distribution of latent variables based on the prior knowledge (i.e. context). Similar with prior network, the recognition network estimates the posterior probability distribution of latent variables based on the posterior knowledge (i.e. context and response). The generation network is used to generate the responses based input context and sampled latent variables. During training, CVAE aims to maximizing the variational lower bound of the conditional log likelihood \citep{StochasticGradientVariationalBayes-kingma-2014,ELBO1-Sohn-2015,ELBO2-Yan-2016}:
\begin{align}
\label{eq:ELBO}
    \nonumber \mathcal{L}(\theta,\phi,\Omega;r,c) &=  \mathbb{E}_{q_{\phi}(z|r,c)}\left[\log p_{\Omega}(r|z, c)\right] \\
    &- \KL(q_{\phi}(z|r,c)||p_{\theta}(z|c)) \, ,
\end{align}
where $\theta,\phi,\Omega$ denote the parameters of the prior network $p_{\theta}(z|c)$, recognition network $q_{\phi}(z|r,c)$ and generation network $p_{\Omega}(r|z, c)$, respectively. $c$ and $r$ represent the context and the response.

\paragraph{Conditional Variational Transformer}
Unlike recurrent neural networks where latent variables are often used as initial decoder states, applying latent variables in transformers is still an open question, and there is currently no general way of combining latent variables with Transformer architectures \citep{DBLP:conf/naacl/HuYLSX22-DELLA}.
Therefore, we review some existing methods of latent variables in transformers.
The construction approaches of latent variables $z$ can be roughly divided into three main categories: (1) initializing a group of parameters as the discrete latent variable \citep{PLATOv1}; (2) pooling of all hidden states of the encoder output and obtain a vector to compute the continuous latent variable, $z \sim \mathcal{N}(\operatorname{Pooling}(h_{\{1,\cdots,n\}}))$, \citep{TCVAE2021-LeFang}; (3) inserting a special token (e.g. $[CLS]$) to compute the sentence-level representation for constructing the latent variable, $z \sim \mathcal{N}(\operatorname{Projection}(h_1^{[CLS]}))$, \citep{GVTandSVT2020,DialogVED-2022-ACL-weichen}.
During generation, existing methods fall into three main paradigms: (1) adding the latent variable to the input embeddings matrix \citep{GVTandSVT2020}; (2) using latent variables as a auxiliary memory to be attended by attention mechanism in each layer \citep{PLATOv1}; (3) adding latent variables with the output decoder states \citep{TCVAE2021-LeFang}.

\subsection{Conditional Hybrid Variational Transformer}
To highlight the respective advantages and to offset disadvantages of continuous latent variables and discrete latent variables, in this section, we propose the Hybrid Latent Variable (HLV).
We describe new methods of modeling these two latent variables, as well as the hybrid of them.
Then we incorporate them into dialogue generation task with a proposed Conditional Hybrid Variational Transformer (CHVT) methods (see Figure~\ref{fig2}).
We specify the details of the training and inference step of the CHVT in the following.

\paragraph{Continuous Latent Variables}
Continuous latent variables are expected to capture more fine-grained diverse features, thereby enhancing the diverse representation of the generated responses.
To achieve this goal, we take inspirations from the outstanding effect of Transformer encoder on Natural Language Understanding tasks, which indicates that the hidden states output by the encoder are rich in semantic information \citep{DBLP:conf/naacl/DevlinCLT19-BERT}.
Combining with the characteristics of Transformer encoder, we propose a new way to construct continuous latent variables, which is different from the aforementioned methods.
We first use the Transformer encoder to encode the input sequence ($\mathbf{x}=x_1, x_2,\cdots, x_n,\cdots,x_{n+m}$) to obtain its final hidden state ($\mathbf{h}=h_1, h_2,\cdots,h_n,\cdots,h_{n+m}$), where, for dialogue generation, $n$ denotes the length of context $c$, $m$ means the length of response $r$.
Then, similar with previous works \citep{VaeTextGeneration-Bowman-2016,kgCVAE-ZhaoTiancheng-2017,CVAE(SPhred)-ShenXiaoyu-2017} that assume the continuous latent variables $z$ follows isotropic Gaussian distribution, we employ fully-connected networks as recognition network $q_{\phi}(z|c,r)\sim \mathcal{N}(\mu,\sigma^2\mathbf{I})$ and prior network $p_{\theta}(z|c)\sim \mathcal{N}(\mu',\sigma'^2\mathbf{I})$:
\begin{align}
\label{eq:two_distribution}
    & \nonumber \begin{pmatrix}
    \mu_1,\cdots,\mu_n \\
    \log(\sigma_1^2),\cdots,\log(\sigma_n^2)
    \end{pmatrix} = \tanh(\begin{pmatrix}
    h_1\\
    \cdots\\
    h_n
    \end{pmatrix} \cdot W_d) \cdot W_u \\
    & \begin{pmatrix}
    \mu_1',\cdots,\mu_n' \\
    \log(\sigma_1'^2),\cdots,\log(\sigma_n'^2)
    \end{pmatrix} = \begin{pmatrix}
    h_1'\\
    \cdots\\
    h_n'
    \end{pmatrix} \cdot W_u' \, ,
\end{align}
where $h_1',\cdots,h_n'$ are hidden states computed by encoder when only the context is input, and $W_{\{d,u\}}, W_u'$ are trainable parameters of recognition network and prior network.
We then obtain $n$ token-level probability distributions for $n$ tokens in the context $c$, every distribution will include the specific features of its corresponding token.
To take full use of these information, we derive an \textit{additive Gaussian mixing} \citep{DBLP:conf/nips/0009SL17-additiveGaussian} to obtain the distribution of sentence-level latent variables $z_s$:
\begin{align}
    p(z_s|c,r) & \sim \mathcal{N}(\sum_{i=1}^nw_i\mu_i, \prod_{i=1}^n\sigma_i^{2w_i}) \label{eq:p(z_s|c,r)}\\
    p(z_s|c) & \sim \mathcal{N}(\sum_{i=1}^nw_i\mu_i', \prod_{i=1}^n\sigma_i'^{2w_i}) \label{eq:p(z_s|c)} \, ,
\end{align}
where $w_i$ denotes the weight of the $i$-th distribution. 
Finally, we use the reparameterization trick \citep{StochasticGradientVariationalBayes-kingma-2014, kgCVAE-ZhaoTiancheng-2017} to obtain samples of $z_s$ either from $p(z_s|c,r)$ (training) or $p(z_s|c)$ (inference).
The sentence-level latent variable $z_s$ will be used for constructing the hybrid latent variable afterwards.

\paragraph{Discrete Latent Variables}
Discrete latent variables are expected to constrain the semantic relevance and coherence between generated responses and the given contexts. 
To this end, we take inspiration from PLATO~\citep{PLATOv1} and self-separated CVAE \citep{Binsun-Sepacvae-ACL2021} to initialize the representations of discrete latent variables and realize the \textit{self-separation training} to optimize the discrete latent variables.
We first create a latent embedding matrix $\LatEmb$ with shape $[K, d_{model}]$, where $K$ denotes the number of discrete latent variables, and $d_{model}$ represents the dimension of each variable.
Afterwards, we use $\LatEmb$ to expand the input dialogue $K$ times but only optimize the sample that reaches the smallest cross-entropy loss:
\begin{align}
\label{eq:self-sepa-training}
    \nonumber \mathcal{J}_{ent} &  = \sum_{i=1}^K\alpha_i \mathcal{J}_i \\
    \text{where} \quad \mathcal{J}_i & = -\log p(r|c, \LatEmb[i])  \\
    \alpha_i & =\left\{
    \begin{array}{ll}
        1 & \operatorname{if}\ \mathcal{J}_i=\min(\mathcal{J}_1,\cdots,\mathcal{J}_K) \\
        0 & \operatorname{otherwise}
    \end{array} \nonumber
    \right.
\end{align}
We assume that a discrete latent variable that is most effective at helping reconstruct the ground truth is the one that best reflects the semantic relationship between response and context.
Therefore, the \textit{self-separation training} will ensure that the discrete latent variable can maintain the relevance and coherence of generated response. 
When inference, we first use $K$ discrete latent variables to generate $K$ responses, then compute the vector representations of them and the given context by Transformer encoder, finally choose the one that has the largest inner-product score with the context.

\paragraph{Hybrid Latent Variables with Transformer}
We propose the Hybrid Latent Variable and integrate it with transformer, namely Conditional Hybrid Variational Transformer, for dialogue response generation.
Since relevance and coherence in dialogue response generation may sometimes be more important than unbridled diversity, the proposed HLV will be dominated by discrete latent variables, enhancing the diversity of generated responses while maintaining the ability in generating relevant and coherent responses.
To tackle this problem, during training, the continuous latent variables $z_s$ of context $c$ will first be sampled from the $p(z_s|c,r)$, and then expanded $K$ times that make it added to the discrete latent variables:
\begin{align}
\label{eq:HLV}
    \HLV = \begin{pmatrix}
    z_s \\
    \cdots \\
    z_s
    \end{pmatrix} + \LatEmb
\end{align}
Then, the decoder of CHVT computes the probability $p(r_i|r_{<i}, c, \HLV)$ based on the contextual memory $\CxtMem$ and $r_{<i}$:
\begin{align}
    p(r|c, \HLV[j]) & = \prod_{i=1}^np(r_i|r_{<i}, c, \HLV[j]) \label{eq:p(r|c,HLV[j])}\\
    p(r_i|r_{<i}, c, \HLV[j]) & = \operatorname{Decoder}(\CxtMem[j], r_{<i}) \\
    \CxtMem[j] & = \begin{pmatrix}
    h_1' \\
    \cdots \\
    h_n'
    \end{pmatrix} + \HLV[j] \, , \quad {j=1,\cdots,K}
\end{align}
Hence, based on the Equation~\eqref{eq:ELBO} and Equation~\eqref{eq:self-sepa-training}, the loss function of CHVT is:
\begin{align}
\label{eq:loss-function}
    \nonumber & \mathcal{J}(\theta, \phi, \psi, \Omega, \LatEmb) = \mathcal{J}_{ent} + \lambda \KL(p(z|c,r)||p(z|c)) \\
    \nonumber & \mathcal{J}_{ent} = \sum_{i=1}^K\alpha_i\mathcal{J}_i, \mathcal{J}_i=-\log(p(r|c, \HLV[i]))\\
    & \alpha_i =\left\{
    \begin{array}{ll}
        1 & if\ \mathcal{J}_i=\min( \mathcal{J}_1,\cdots,\mathcal{J}_K ) \\
        0 & otherwise,
    \end{array} 
    \right.
\end{align}
where $\theta, \phi, \psi, \Omega, \LatEmb$ are parameters of CHVT, and $\lambda$ is the scale factor of KL divergence.

\paragraph{Inference Phase}
During inference, CHVT use the prior distribution $p(z_s|c)$ to sample the sentence-level continuous latent variable $z_s$ and mix $z_s$ with discrete latent variables to construct HLV.
Based on the Equation~\eqref{eq:p(r|c,HLV[j])}, CHVT generate $K$ responses for the same context.
To figure out which response is the best for the context, we first obtain the vector representations of the context and responses by using the encoder of CHVT . And then we compute the inner-product score between each generated response vector and context vector.
Finally, we choose the response with the largest inner-product score.

\subsection{Theoretical Results\footnote{All detailed proofs will be available in the final writing.}}


In dialogue response generation, a vanilla conditional variational auto-encoder (CVAE) inevitably needs to model the one-to-many phenomenon, i.e. one context $c$ corresponds to $n$ responses ($r_1, r_2,\cdots,r_n$, $n \geq 2$). Thence, the prior distribution $p(z|c)$ will be forced to align with multiple posterior distributions ($p(z|c,r_1), p(z|c,r_2),\cdots, p(z|c,r_n)$) during training. We consider the optimal KL-divergences for any given one-to-many samples.
\begin{theorem}
\label{theorem_one}
For one-to-many phenomenon existed, the sum of $\KL$ part in Equation~\eqref{eq:loss-function} has a extreme minimum $\xi \geq 0:$
\begin{align}
    \nonumber \xi & = \sum_{i=1}^n\KL(p(z|c,r_i)||p(z|c)) \\
    & =\frac{n}{2}\log(1+\frac{\sum_{i=1}^n(\mu_i -\bar{\mu})^2}{n}\frac{1}{\bar{\sigma}^2}) \, ,
\end{align}
where $\bar{\mu}=\frac{1}{n}\sum_{i=1}^n\mu_i$, and $\bar{\sigma}=\frac{1}{n}\sum_{i=1}^n\sigma_i$.
\end{theorem}

Note that $\xi \propto \frac{\sum_{i=1}^n(\mu_i-\bar{\mu})}{n}$ and $\xi \propto \frac{1}{\bar{\sigma}^2}$. 
The $\frac{\sum_{i=1}^n(\mu_i-\mu^*)}{n}$ can be regarded as the variance of these ($\mu_1,\mu_2,\cdots,\mu_n$) of posterior distributions, which means that the larger the gap between the responses corresponding to the context, the larger the corresponding minimum value of KL divergence.

These results explained another possible reason of the posterior vanishing problem in the dialogue response generation task: According to the principle of conditional variational mechanism, $\mu$ of posterior distribution is expected to capture the implicit representation of the response conditional on the given context, while $\sigma$ of posterior distribution is expected to be smaller enough to ensure the stability of the sampling process.
However, due to the one-to-many problem, the better $\mu$ and $\sigma$ of the posterior distribution are learned, the larger and lower bound of the KL divergence.
And the objective function expects the KL divergence to be reduced to 0, which tends the models to selects meaningless $\mu$ and $\sigma$, that is, invalidates the posterior distribution.

Existing efforts usually use \textsc{KL annealing} \citep{VaeTextGeneration-Bowman-2016}, \textsc{Bag-of-Words} (BOW) loss \citep{kgCVAE-ZhaoTiancheng-2017, GVTandSVT2020} and \textsc{KL thresholding} \citep{DBLP:conf/acl/ZhuBLMLW20-BNVAE} to address the latent vanishing problem.
The \textsc{KL annealing} trick introduces KL divergence with a weight from 0 to 1 during first $k_{ann}$ steps.
Based on the \textit{Theorem~\ref{theorem_one}}, large $k_{ann}$ may give better results, as it weakens the $\KL$ part of objective function and fully uses of the latent variables sampled by posterior distribution.
We observed that BOW loss may sometimes fail, as we noticed that BOW loss will increase the discrimination between latent variables, i.e., it increase the distance among different $\mu$, thus increasing $\xi$ and interferes with the training of the model.
As for \textsc{KL thresholding}, $\xi$ can be a dynamic thresholding of KL divergence, but it is a challenge to estimate $\xi$ during training stage.
Therefore, to train the CHVT, we only introduce the KL annealing trick with a large warm-up batch size.

In addition to the above findings, we also have:
\begin{proposition}
\label{proposition_one}
Based on Theorem~\ref{theorem_one}, if $\forall i \in [1,n], \mu_i\geq 0$, then there is an upper bound $\eta$ for $\xi$ that satisfies $\xi \leq \eta$.
\begin{align}
    \eta = \frac{n}{2}\log(1+(n-1)(\frac{\bar{\mu}}{\bar{\sigma}})^2)
\end{align}
\end{proposition}
Based on the \textit{Theorem~\ref{theorem_one}} and \textit{Proposition~\ref{proposition_one}}, we propose a relaxed KL divergence ($\RKL$) to smoothly train the CHVT:
\begin{align}
    \RKL = \max(\KL - \frac{\eta}{n}, 0),
\end{align}
which aims to make the average KL divergence only need to be lower than $\frac{\eta}{n}$, not to 0, for one-to-many samples.
By relaxing the KL divergence, the contradiction, that is the objective of conditional variational model expects KL decrease to 0 but the one-to-many problem causes KL greater than 0, will be weaken, so as to avoid disappearance of a posterior vanishing problem.
For ease of calculation, we use the expectation of $\mu$ to approximate $\bar{\mu} \approx \mathbb{E}(\mu_1,\mu_2,\cdots,\mu_n) \approx \frac{1}{b}\sum_{i=1}^b\mu_i$, where $b$ denotes the batch size.

Through theoretical analysis, we introduce several training methods for CHVT, i.e. elimination of the BOW loss, large $k_{ann}$ step of KL annealing, and relaxed KL divergence.

\begin{table}[t]
\centering
\small
\renewcommand\tabcolsep{3.8pt}
\begin{tabular}{lcccc}
\toprule
            & \# Vocab  & \# Train & \# Valid & \# Test \\ \midrule
    \texttt{DailyDialog}   & 17,930  & 68,066 & 6,820 & 6,841 \\
    \texttt{OpenSubtitles} & 21,177  &  200K  &  20K & 10K   \\
\bottomrule
\end{tabular}
\caption{Statistics of the datasets used for experiments.}
\label{tab:data_statistics}
\end{table}

\section{Experiment}
\label{sec:experiment}
\subsection{Data Setting}

We evaluate our methods over two well-established open-domain dialogue datasets: \texttt{DailyDialog}~\citep{dailydialog2017} and \texttt{Opensubtitles}~\citep{opensubtitles2016}. 
For our purpose, we pre-process two open domain dialogue corpora.
From a multi-turn dialogue $(u_1,u_2,...,u_T)$, we can extract $T-1$ single-turn dialogues $[(u_1,u_2),(u_2,u_3),...,(u_{T-1},u_{T})]$, where $u$ represents an utterance.
We collected all dialogue pairs, reduced the repeat pairs, and divided them into training, validation and test sets.
Table~\ref{tab:data_statistics} lists key statistics of our datasets.

\begin{table*}[t]
\centering
\small
\begin{tabular}{lc|ccc|c|cccc|cc}
\toprule
\textit{DailyDialog}  & PPL $\downarrow$   & \multicolumn{3}{c|}{Disntinct-1/2/3}        & Len. & \multicolumn{4}{c|}{BLEU-1/2/3/4}                             & EA.         & Cohe.       \\
\midrule
TRF             & 37.20          & 0.0103          & 0.0465          & 0.0934          & 10.07           & 0.2966          & 0.2356          & 0.1949          & 0.1543          & 0.8023          & 0.6838          \\
GVT             & 38.05          & 0.0059          & 0.0186          & 0.0299          & 10.51           & 0.2816          & 0.2239          & 0.1856          & 0.1472          & 0.7864          & 0.6662          \\
SVT             & 36.91          & 0.0113          & 0.0419          & 0.0779          & 7.95            & 0.2631          & 0.2079          & 0.1713          & 0.1353          & 0.7703          & 0.6360          \\
TCVAE           & 40.15          & 0.0160 & 0.0701          & 0.1474          & 7.78            & 0.2786          & 0.2181          & 0.1791          & 0.1412          & 0.8031          & 0.6881          \\
CHVT                        & 21.36    & 0.0156   & 0.1093    & 0.3074                 & 11.03            & 0.3472             & 0.2773            & 0.2317            & 0.1847            & \textbf{0.8365}& \textbf{0.7683}            \\
CHVT + $\RKL$                  & \textbf{18.92}           & \textbf{0.0161}                  & \textbf{0.1304}                 & \textbf{0.3918}                 & \textbf{11.75}            & \textbf{0.3496}             & \textbf{0.2844}            & \textbf{0.2396}            & \textbf{0.1917}            & 0.8305                        & 0.7608                     \\
\midrule
PLATO           & 12.73          & \textbf{0.0620}  & 0.2841          & 0.4993          & 8.31            & 0.3181          & 0.2485          & 0.2042          & 0.1612          & 0.8277          & 0.7916          \\
PLATO + HLV & \textbf{6.17}  & 0.0523          & \textbf{0.3170} & \textbf{0.6344} & \textbf{10.78} & \textbf{0.3720} & \textbf{0.2962} & \textbf{0.2469} & \textbf{0.1966} & \textbf{0.8449} & \textbf{0.8123}\\
BART-base       & 11.59       & 0.0764          & 0.2442          & 0.4099          & 6.35            & 0.0997          & 0.0801          & 0.0669          & 0.0533          & 0.5968          & 0.5842          \\
BART + HLV       & \textbf{6.28}  & \textbf{0.0773}  & \textbf{0.3580} & \textbf{0.6903}                          & \textbf{9.87}             & \textbf{0.1758}             & \textbf{0.1455}            & \textbf{0.1238}            & \textbf{0.0996}            & \textbf{0.6607}               & \textbf{0.6498} \\

\midrule
\textit{OpenSubtitles} & PPL $\downarrow$   & \multicolumn{3}{c|}{Disntinct-1/2/3}        & Len. & \multicolumn{4}{c|}{BLEU-1/2/3/4}                             & EA.         & Cohe.       \\
\midrule
TRF             & 49.11          & 0.0020          & 0.0068          & 0.0116          & 7.10            & 0.1902          & 0.1478          & 0.1205          & 0.0947          & 0.8014          & 0.6983          \\
GVT             & 44.32          & \textbf{0.0049} & 0.0246          & 0.0587          & 9.62   & 0.2677          & 0.2123          & 0.1758          & 0.1393          & 0.8156          & 0.7251          \\
SVT             & 49.39          & 0.0020          & 0.0066          & 0.0115          & 4.93            & 0.1478          & 0.1125          & 0.0902          & 0.0701          & 0.7959          & 0.6741          \\
TCVAE           & 51.54          & 0.0027          & 0.0092          & 0.0165          & 8.34            & 0.2162          & 0.1705          & 0.1406          & 0.1112          & 0.7890          & 0.6619          \\
CHVT            & \textbf{27.09}    & 0.0045    & \textbf{0.0450}       & \textbf{0.1906}   & 10.87            & 0.3759             & 0.3013            & 0.2516            & 0.2002            & 0.8441               & \textbf{0.7793}            \\
CHVT + $\RKL$          & 29.24                    & 0.0035                           & 0.0330                          & 0.1516                          & \textbf{11.96}            & \textbf{0.3989}             & \textbf{0.3230}            & \textbf{0.2711}            & \textbf{0.2164}            & \textbf{0.8483}               & 0.7757                     \\
\midrule
PLATO  & 24.25          & \textbf{0.0606} & 0.2555 & 0.4782 & 10.83  & 0.3581 & 0.2885 & 0.2414 & 0.1926 & 0.8159          & \textbf{0.8639} \\
PLATO + HLV & \textbf{5.46}                      & 0.0391          & \textbf{0.2561} & \textbf{0.6257} & \textbf{13.30} & \textbf{0.4465} & \textbf{0.3622} & \textbf{0.3046} & \textbf{0.2437} & \textbf{0.8606} & 0.8589        \\
BART-base       & 21.84          & 0.0717          & 0.2392          & 0.3930          & 7.51            & 0.1024          & 0.0837          & 0.0706          & 0.0565          & 0.5275          & 0.5362          \\
BART + HLV       & \textbf{4.51}  & \textbf{0.0963} & \textbf{0.4120} & \textbf{0.7299} & \textbf{10.00}  & \textbf{0.1574} & \textbf{0.1312} & \textbf{0.1118} & \textbf{0.0901} & \textbf{0.6454} & \textbf{0.6474} \\
\bottomrule
\end{tabular}
\caption{Auto-evaluation results over the test sets of \texttt{DailyDialog} and \texttt{OpenSubtitles} datasets. ``CHVT'' and ``+HLV'' indicate our hybrid latent variables models.}
\label{main_test_result}
\end{table*}

\subsection{Baselines}
\label{sect:dia-baselines}
\paragraph{Non-Pretrained Baselines} We compare our Conditional Hybrid Variational Transformer with state-of-the-art Transformer-based generative models: a basic Transformer encoder-decoder model (TRF) \citep{Transformer-Vaswani-2017}, a Global Varitional Transformer (GVT) and a Sequential Variational Transformer (SVT) \citep{GVTandSVT2020}, a Transformer Conditional Variational AutoEncoder (TCVAE) \citep{TCVAE2021-LeFang}.
We train them on \texttt{DailyDialog} and \texttt{OpenSubtitles} from scratch.

\paragraph{Pre-trained Baselines} To explore the generalization of HLV in pre-trained dialogue models, we adapt HLV over PLATO-v1 (132M parameters; \citet{PLATOv1}), a pre-trained dialogue generation model based on UniLM \cite{dong2019unified} with only discrete latent variable,\footnote{Thus we only add continuous latent variable to PLATO.} and BART-base~(110M parameters; \citet{DBLP:conf/acl/LewisLGGMLSZ20-BART}), a pre-trained sequence-to-sequence transformer model without latent variables.
We finetune these two models with HLV, noted as ``PLATO + HLV'' and ``BART + HLV'', over our training sets.





\subsection{Automatic Evaluation Results}

Table~\ref{main_test_result} reports the automatic results over the test sets of \texttt{DailyDialog} and \texttt{OpenSubtitles}.

\paragraph{Comparison with Non-Pretrained Baselines}
Compared with non-pretrained baselines (i.e. TRF, GVT, SVT and TCVAE), our CHVT/CHVT+$\RKL$ achieves better performance in terms of most metrics.
Specifically, our CHVT/CHVT+$\RKL$ achieves the best Distinct-2/3, Len., BLEU-1/2/3/4, EA. and Cohe. scores on both datasets, which demonstrates the superior performance of our model on generating diverse, coherent and relevant responses.

It can be observed that applying $\RKL$, rather than $\KL$, increases the BLEU scores, demonstrating that latent variables trained by $\RKL$ better capture the features that can reconstruct the response. This also shows that $\RKL$ can help tackle the one-to-many problem for the training of latent variables to some extent, so that latent variables can capture more information for reconstructing ground-truth.

We also observe that (1) GVT achieve a better result on \texttt{OpenSubtitles} but poor result on \texttt{DailyDialog}; (2) SVT performs poor results on both two datasets; (3) TCVAE achieves decent performance on most metrics, but a high PPL score.
These phenomena indicate the difficulty of training latent variable models: (1) With a small amount of data in \texttt{DailyDialog}, GVT performed weaker than TRF, but the results of \texttt{OpenSubtitles} were opposite, indicating that GVT may require strict prerequisites, i.e., more data for the model to learn a well representation of $[CLS]$.
(2) SVT uses predicted tokens to summarize $z$, which may be affected by high frequency tokens, limiting the generation of better responses.
(3) TCVAE constructs $z$ based all hidden states of the encoder and strengthens $z$ by adding it to the input embedding, inserting it into self-attention layer, and adding it to the output hidden state.
Although these settings improve the overall results, higher PPL suggests that SVT may be difficult to generalize well to other datasets.


\begin{table*}[t]
\centering
\small
\begin{tabular}{llc|ccc|c|cccc|cc}
\toprule
&  & PPL $\downarrow$   & \multicolumn{3}{c|}{Disntinct-1/2/3}        & Len. & \multicolumn{4}{c|}{BLEU-1/2/3/4}                             & EA.         & Cohe.       \\
\midrule
\multirow{3}{*}{\textit{Validation}} & CHVT                        & \textbf{21.21}           & \textbf{0.0146}                           & \textbf{0.1030}                 & \textbf{0.2911}                 & \textbf{11.40}            & \textbf{0.3578}             & \textbf{0.2864}            & \textbf{0.2397}            & \textbf{0.1913}            & \textbf{0.8369}               & \textbf{0.7527}            \\
 & \quad w/o. CLV & 35.98                              & 0.0093          & 0.0456          & 0.0969          & 10.66          & 0.3540          & 0.2782          & 0.2288          & 0.1804          & 0.8233          & 0.7008          \\
 & \quad w/o. DLV & \textbf{18.50} & 0.0012          & 0.0137          & 0.0278          & \textbf{33.42} & 0.1560          & 0.1394          & 0.1228          & 0.1006          & 0.5736          & 0.5944          \\
\midrule
\multirow{3}{*}{\textit{Test}} & CHVT                        & \textbf{21.36}           & \textbf{0.0156}                           & \textbf{0.1093}                 & \textbf{0.3074}                 & \textbf{11.03}            & \textbf{0.3472}             & \textbf{0.2773}            & \textbf{0.2317}            & \textbf{0.1847}            & \textbf{0.8365}               & \textbf{0.7683}            \\
& \quad w/o. CLV & 36.05                              & 0.0098          & 0.0500          & 0.1105          & 11.00          & 0.3231          & 0.2581          & 0.2151          & 0.1711          & 0.8151          & 0.7319          \\
& \quad w/o. DLV & \textbf{18.46} & 0.0011          & 0.0137          & 0.0277          & \textbf{33.24} & 0.1599          & 0.1430          & 0.1260          & 0.1033          & 0.5744          & 0.5883 \\
\bottomrule
\end{tabular}
\caption{Ablation results over the validation and test sets of \texttt{DailyDialog}.}
\label{tab:ablation_results}
\end{table*}

\paragraph{Comparison with Pre-trained Baselines}
We compare the results of pre-trained baselines with fine-tuning using our proposed HLV.
Since PLATO has pre-trained with discrete latent variables, we only analysis the utility of continuous latent variables.
In order to not destroy the knowledge learned from PLATO pre-training, 
we construct HLV by adding the continuous latent variables to the original discrete latent variables of PLATO, and using the HLV instead of the discrete latent variables in PLATO for training.

Table~\ref{main_test_result} shows that even for a well-pretrained dialogue model with discrete hidden variables, such as PLATO, simply introducing HLV can still improves the performance with a large margin.
Also for BART, which does not involve any types of latent variables in the pretraining phase, we simply incorporate HLV into it in the same way as CHVT.
The results show that even if the latent variables are not exposed during pre-training, introducing HLV during fine-tuning can significantly improve the model performance for dialogue generation.

In a nutshell, these results shows the ability of HLV and CHVT to generate diverse, relevant and coherent dialogue responses across different model setups.

\subsection{Ablation Study}
\label{sec:ablation}

Table~\ref{tab:ablation_results} reports the results of the ablation study of two latent variables.
After removing the continuous latent variable (CLV) and the discrete latent variable (DLV), respectively, most of the metric scores decreased.
we can observe that after removing CLV, the Distinct-1/2/3 of the model on validation and test sets of \texttt{DailyDialog} are reduced by 36.02\% / 55.72\% / 66.73\% and 37.12\% / 54.28\% / 64.04\%, respectively. However, the BLEU-1/2/3/4, EA. and Cohe. on validation and test sets only dropped an average of 3.78\% and an average of 5.94\%, respectively.
This indicates that CLV has a greater impact on diversity metric.
Furthermore, we also observe that the performance decreased the most when removing DLV, indicating that DLV is more important for this task.
The overall result is also in line with our expectation, as stated before, that to prevent the risk of uncontrollable diversity, our HLV is mainly dominated by DLV, and CLV is responsible for providing diverse features that are difficult to capture by DLV.

These findings demonstrate the contribution of discrete latent variables in facilitating the generation of relevant and coherent responses.

\subsection{Human Evaluation}
Table~\ref{tab:human_eval} reports the result of human evaluation. Following the work of \citet{AdverREGS-LiJiwei-2017,Binsun-Sepacvae-ACL2021}, We hire three annotators to do preliminary human evaluations by comparing 50 responses generated by PLAO, PLATO+HLV, CHVT and TCVAE w.r.t. Diversity (Div.), Relevance (Rel.) and Fluency (Flu.). We notice that our CHVT achieves a significant performance on diversity and relevance. Also, PLATO achieves a significant improvement on diversity by using HLV. These preliminary results could strength our statements.

\begin{table}[t]
\centering
\small
\begin{tabular}{lccc}
\toprule
 & CHVT vs. TCVAE   & PLATO+HLV vs. PLATO &   \\
\midrule
Div.  & 56.67 / 12.0 [0.674] & 39.33 / 18.00 [0.663] \\
Rel.  & 30.67 / 20.0 [0.528] & 25.33 / 20.00 [0.557] \\
Flu.  & 24.67 / 24.0 [0.658] & 18.67 / 17.33 [0.490] \\
\bottomrule
\end{tabular}
\caption{The win/loss(\%)[kappa] on Diversity, Relevance and Fluency over the test set of \texttt{DailyDialog}.}
\label{tab:human_eval}
\end{table}

\subsection{Effective Study}

\begin{figure}[t]
\centering
\includegraphics[width=0.45\textwidth]{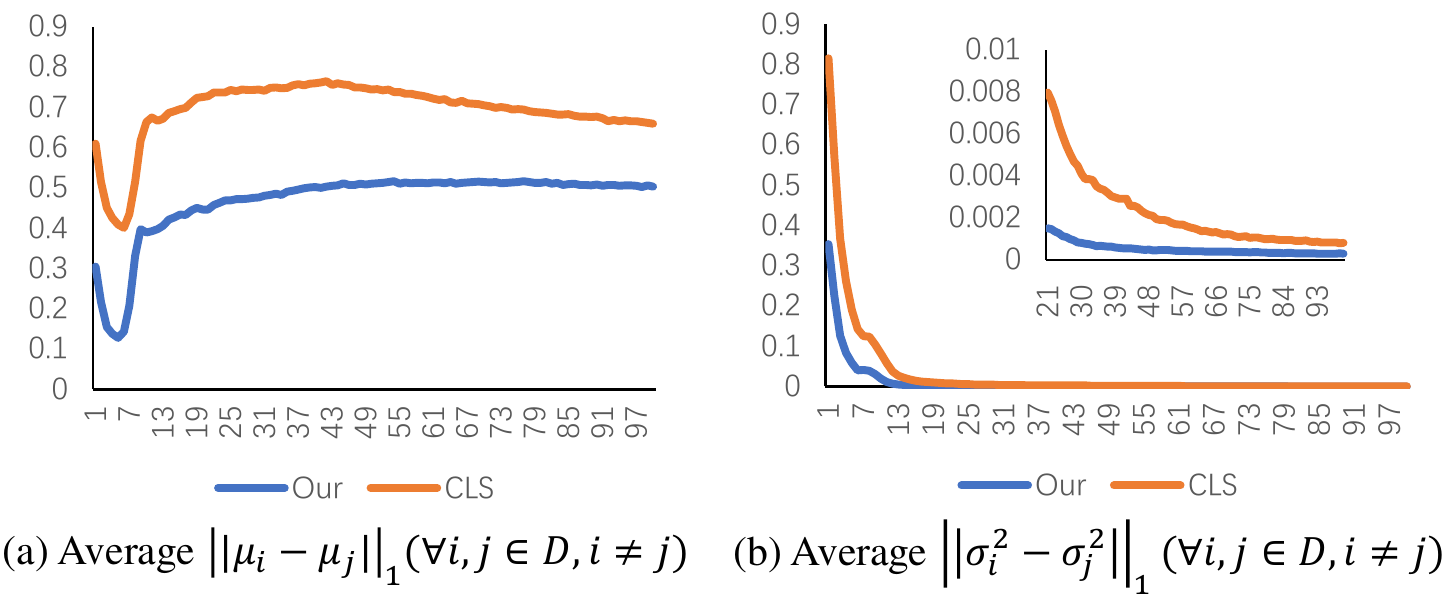} 
\caption{The averaged $L1$ distance of all pairs of two posterior Gaussian distribution.}
\label{fig:effective_analysis}
\end{figure}

To validate our \textit{Theorem~\ref{theorem_one}}, the effectiveness of constructing continuous latent variables, we conduct an analysis experiment.
We first extract $2,000$ dialogue pairs $D$ from the training set of \texttt{DailyDialog} to train a Transformer CVAE.
This CVAE only needs to reconstruct the responses based on the encoder output and posterior continuous latent variables.
During training, we counted the $L1$ distance between $\mu_i, \mu_j$ (and $\sigma_i^2, \sigma_j^2$) of any two posterior distributions ($p_i \sim \mathcal{N}$($\mu_i, \sigma_i^2$), $p_j \sim \mathcal{N}$($\mu_j, \sigma_j^2$)) after each epoch.

The results are shown in Figure~\ref{fig:effective_analysis}.
``CLS'' denotes that using $[CLS]$ token to construct the continuous latent variables.
We observe that $\sigma^2$ will be close to each other during training, which satisfies the assumption for \textit{Theorem~\ref{theorem_one}} and validates the accuracy of \textit{Theorem~\ref{theorem_one}}.
However, the averaged distance of $\mu$ among posterior distributions also demonstrates that directly using the averaged $\mu$ of all posterior distributions in one batch to approximate the $\frac{1}{n}\sum_{i=1}^n$ may fail.
Therefore, how to apply \textit{Proposition 1} to effectively train conditional variational mechanisms in dialogue response generation will be a challenge for our future work.

\section{Conclusion}
This paper proposes the Hybrid Latent Variable to make full use of advantages of the continuous and discrete latent variables.
We also propose a Conditional Hybrid Variational Transformer to construct and incorporate HLV into dialogue response generation.
To better train CHVT and HLV, we theoretically analyze why vanilla conditional variational mechanisms are hard to train stably in dialogue generation tasks, and give suggestions for stabilizing the training of latent variables.
In addition, we also propose several techniques for stabilizing the optimization procedure, including removing the BOW loss, a KL annealing trick, and a relaxed KL divergence loss.
We conduct extensive experiments to show the superior performance of the proposed HLV and CHVT in dealing with the dialogue response generation task.
Experimental results show that CHVT is superior to non-pretrained baselines w.r.t. diversity, relevance and coherence metrics. Furthermore, the experiments with pre-trained baselines demonstrate the compatibility and advantages of HLV during finetuning, that incorporating HLV enables better overall performance on dialogue response response generation task.

\section*{Acknowledgments}
We would like to thank the anonymous reviewers for their constructive comments. This research is supported by Beijing Natural Science Foundation (No.4222037 and L181010). Kan Li is the corresponding author.
\bibliography{aaai23}

\end{document}